\documentclass[letterpaper, 10 pt, journal, twoside]{ieeetran}

\IEEEoverridecommandlockouts

\usepackage{graphics}
\usepackage{graphicx}
\usepackage{epsfig}
\usepackage{mathptmx}
\usepackage{times}
\usepackage{amsmath}
\usepackage{bm}
\usepackage{amssymb}
\usepackage{cite}
\usepackage{multirow}
\usepackage{booktabs}
\usepackage{makecell}
\usepackage[table]{xcolor}
\usepackage{pgfplots}
\usepackage[export]{adjustbox}
\usepackage{afterpage}
\usepackage{xurl}
\usepackage{hyperref}
\usepackage{subcaption}
\usepackage[normalem]{ulem}
\usepackage{soul}
\usepackage{svg}
\usepackage{algorithm}
\usepackage{algorithmicx}
\usepackage{algpseudocode}

\pgfplotsset{compat=1.17}

\definecolor{invbg}{HTML}{FFF2F4}
\definecolor{randbg}{HTML}{F2F7FF}
\definecolor{soothingblue}{RGB}{40, 120, 200}
\definecolor{softred}{RGB}{200, 80, 80}

\newcommand{\stabcell}[2]{%
  \ifcase#1 \cellcolor{red!12}{#2}%
  \or       \cellcolor{orange!12}{#2}%
  \or       \cellcolor{yellow!25}{#2}%
  \or       \cellcolor{yellow!35}{#2}%
  \or       \cellcolor{green!10}{#2}%
  \else     \cellcolor{green!30}{#2}%
  \fi
}

\newcommand{\added}[1]{#1}
\newcommand{\deleted}[1]{\unskip}

\makeatletter
\newcommand{\StateHL}[2][invbg]{%
  \State\begingroup
    \setlength{\fboxsep}{1pt}%
    \colorbox{#1}{%
      \parbox{\dimexpr\linewidth-\ALG@thistlm-2\fboxsep}{#2}%
    }%
  \endgroup
}
\makeatother

\title{ShelfAware: Real-Time Visual-Inertial Semantic Localization in Quasi-Static Environments with Low-Cost Sensors}

\author{Shivendra Agrawal, Jake Brawer, Ashutosh Naik, Alessandro Roncone, and Bradley Hayes%
\thanks{Manuscript received: December 9, 2025; Revised: March 4, 2026; Accepted: March 30, 2026.}%
\thanks{This paper was recommended for publication by Editor Ayoung Kim upon evaluation of the Associate Editor and Reviewers' comments.}%
\thanks{The authors are with the Department of Computer Science, University of Colorado Boulder, Boulder, CO 80309 USA (e-mail: shivendra.agrawal@colorado.edu).}%
\thanks{Digital Object Identifier (DOI): 10.1109/LRA.2026.3682613.}%
}

\begin{document}

\maketitle

\begin{center}
\small
This is the accepted author manuscript of an article accepted for publication in
IEEE Robotics and Automation Letters. The final version of record is available at:
\href{https://doi.org/10.1109/LRA.2026.3682613}{https://doi.org/10.1109/LRA.2026.3682613}.
\end{center}


\begin{abstract}
Many indoor workspaces are \emph{quasi-static}: their global geometric layout is stable, but local semantics change continually, producing repetitive geometry, dynamic clutter, and perceptual noise that defeat standard vision-based localization. We present \emph{ShelfAware}, a semantic particle filter for robust global localization that treats scene semantics as \emph{statistical evidence over object categories} rather than fixed quantity landmarks. ShelfAware fuses a depth likelihood with a category-centric semantic similarity and uses a precomputed bank of semantic viewpoints to perform \emph{inverse semantic proposals} inside Monte Carlo Localization (MCL), yielding fast, targeted hypothesis generation on low-cost, vision-only hardware. To demonstrate perception-agnostic scalability, we evaluate ShelfAware across two domains. In a rigorously controlled mock retail environment, ShelfAware achieves a 97\% global localization success rate, maintaining the highest tracking success (66\%) across cart, wearable, and dynamic occlusion conditions. Furthermore, in a 3,500 sq. ft. operational grocery store leveraging an open-vocabulary vision pipeline, ShelfAware significantly outperforms both geometric and fixed-quantity semantic baselines. By modeling semantics distributionally and leveraging inverse proposals, ShelfAware resolves geometric aliasing, providing an infrastructure-free building block for mobile and assistive robots in dynamic real-world environments.
\end{abstract}



\section{INTRODUCTION}

\IEEEPARstart{M}{any} real-world indoor environments, such as retail stores and warehouses, are \emph{quasi-static}: their global geometric layout (e.g., walls, shelving units) remains permanent over long horizons, but their local semantic contents change continually. In these settings, traditional geometry-based particle filters like Adaptive Monte Carlo Localization (AMCL) \cite{foxKLDSamplingAdaptiveParticle2001} suffer from severe perceptual aliasing. The combination of visually repetitive aisle geometry, dynamic clutter, and the noisy observations typical of low-cost wearable or mobile sensors routinely violates static map assumptions and causes rapid particle impoverishment \cite{gharpureRobotassistedShoppingBlind2008}. Consequently, robust, start-anytime global localization in these GPS-denied environments remains a critical open problem for autonomous and assistive systems.

\begin{figure}[t]
  \centering
 \includegraphics[width=\columnwidth]{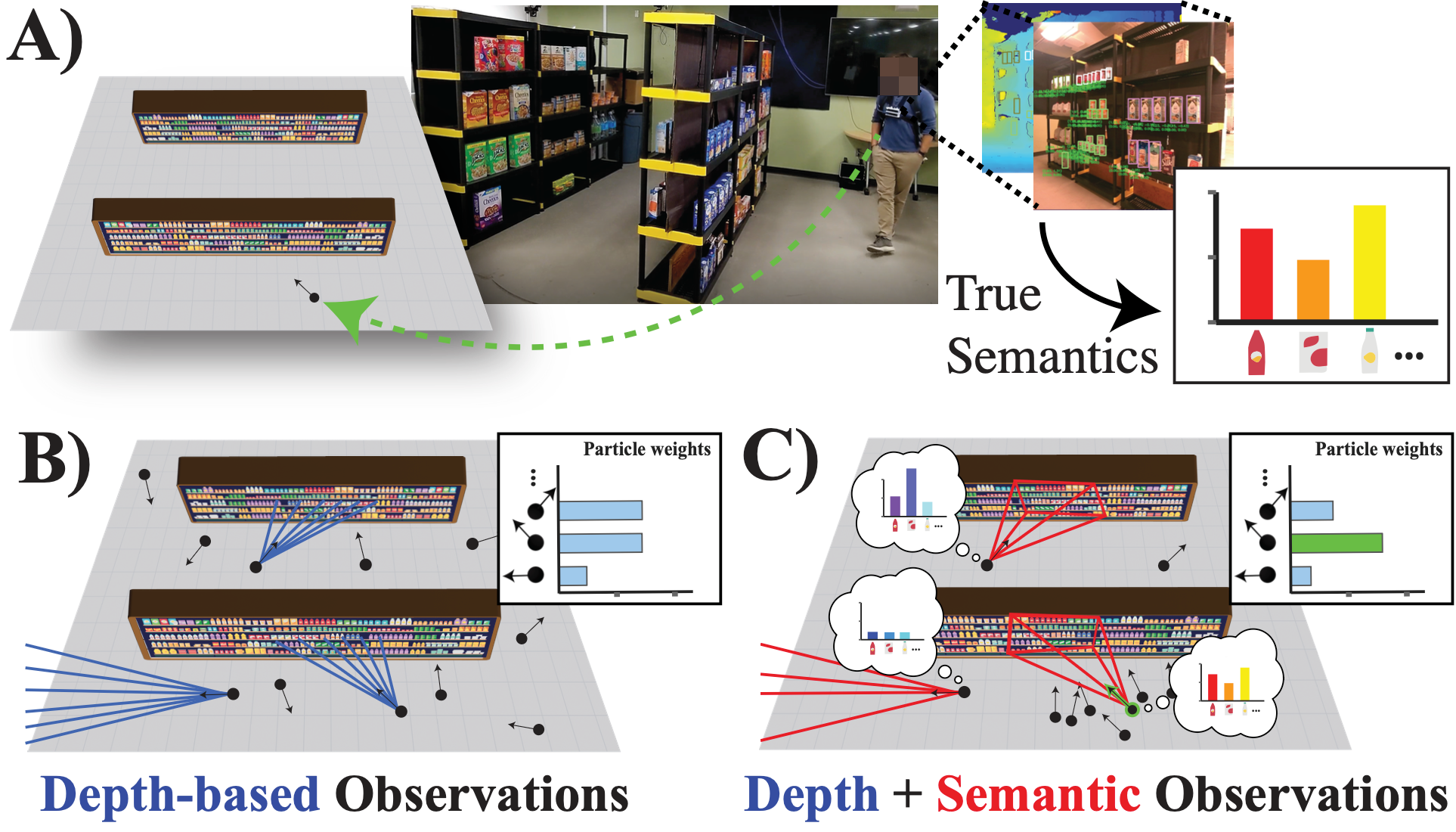}
  \caption{An overview of ShelfAware. A) A retail-like environment. B) Depth-based observation models in particle filtering rely solely on geometric features, which are ambiguous in long, repetitive aisles and lead to weak particle discrimination. C) ShelfAware injects particles based on semantic cues, enabling more distinctive and robust particle weighting combined with the depth observation model and improved global localization in retail-like environments.}
  \label{fig:overview}
\end{figure}

To address these challenges, we present ShelfAware (Fig. \ref{fig:overview}), a semantic particle filter tailored for robust global localization in quasi-static environments. Unlike prior approaches that treat semantics as fixed, discrete landmarks \cite{zimmermanRobustOnboardLocalization2022, zimmermanLongTermLocalizationUsing2023, goswamiEfficientRealTimeLocalization2023a, xieRobustLifelongIndoor2024}, ShelfAware models semantic features as probabilistic distributions over object counts and spatial arrangements. This representation captures the intrinsic variability of real-world environments, preserving the statistical structure necessary for stable observation-driven localization even as local object configurations continuously fluctuate.

We evaluate ShelfAware across two domains: a semantically dense mock retail environment enabling rigorous motion-capture ground truth, and a 3,500 sq. ft. operational grocery store demonstrating open-world scalability. Experiments on consumer-grade wearable and cart-mounted sensors highlight the system's robustness for practical deployment across both autonomous service robots and assistive devices. 

Our core technical contributions are twofold:
\begin{itemize}
    \item A semantic mapping and observation design that encodes object collections as statistical distributions over counts and arrangement, providing inherent robustness to semantic perturbations and flux. 
    \item A real-time, inverse observation model-based particle filter that leverages this representation on low-cost, vision-based hardware to efficiently propose and evaluate high-quality global pose hypotheses.
\end{itemize}

\section{RELATED WORK}
\subsection{Vision and Semantics-Based Localization in Quasi-Static Environments}
Particle filter-based localization methods such as Monte Carlo Localization (MCL) and its adaptive variant AMCL \cite{foxKLDSamplingAdaptiveParticle2001} remain standard for onboard robot localization due to their scalability and integration with modern navigation frameworks \cite{nav2_amcl}. However, their reliance on static geometric maps and feature-rich depth data makes them brittle in quasi-static or dynamic environments where geometry is repetitive or transient. This limitation has driven research toward semantic localization, where environmental understanding extends beyond geometry to include object-level cues and contextual information \cite{yinSurveyGlobalLiDAR2024}.

Earlier work in semantic localization augments geometric maps with explicit, object- or region-level labels that serve as long-term landmarks for place recognition and drift reduction \cite{zimmermanLongTermLocalizationUsing2023, goswamiEfficientRealTimeLocalization2023a, xieRobustLifelongIndoor2024}. Text-based cues, such as signage or packaging, have also proven effective as distinctive features in structured indoor spaces \cite{zimmermanRobustOnboardLocalization2022}. More recent efforts adopt implicit semantic representations, encoding spatial and semantic structure jointly in learned neural fields \cite{kuangIRMCLImplicitRepresentationBased2023}, enabling continuous observation models compatible with particle filtering.
Yet, both explicit and implicit methods often assume stable semantic identities, assumptions that rarely hold in quasi-static domains like warehouses or retail stores, where local semantics fluctuate even when global geometry remains constant.

Despite these advances, few approaches explicitly address semantic volatility: the gradual yet continual change in object distributions and arrangements that characterizes quasi-static environments. Traditional semantic-SLAM systems \cite{goswamiEfficientRealTimeLocalization2023a, crabbLightweightApproachLocalization2023} and semantic visual positioning frameworks \cite{ruiMultiSensoryBlindGuidance2021, chenCanWeUnify2020} demonstrate the promise of integrating semantics into localization, but their models typically treat detected landmarks as fixed or sparsely varying entities. This mismatch between model assumptions and real-world semantic drift leads to degraded performance in settings with restocking, occlusions, or partial observability.
\added{Recent deep visual localization methods \cite{loiseau2025alligat0r, dong2025reloc3r} demonstrate impressive accuracy but require server-grade compute, precluding real-time use on wearable edge devices. Furthermore, while modern visual SLAM \cite{al2024review} excels at local tracking, it typically requires extensive exploration to achieve global localization against a prior map, limiting the instantaneous recovery needed for on-demand assistance.}
\subsection{Applications in Assistive and Human-Centered Robotics}

Quasi-static indoor environments such as retail stores also appear prominently in assistive and human-centered robotics. Many assistive navigation systems for people with visual impairments depend on reliable global localization to support guidance, object retrieval, or wayfinding \cite{kulyukinRoboCartRobotassistedNavigation2005,lopez-de-ipinaBlindShoppingEnablingAccessible2011,bigham2010vizwiz}. However, prior systems often sidestep this challenge, relying instead on environmental instrumentation (e.g., RFID tags \cite{lopez-de-ipinaBlindShoppingEnablingAccessible2011}, Bluetooth beacons) or assuming that the user is already localized within a specific aisle or region \cite{agrawalShelfHelpEmpoweringHumans2023}. Recent conversational and multimodal assistance frameworks \cite{kamikubo2024we,kaniwa2024chitchatguide} have improved interaction but still rely on external positioning aids. ShelfAware complements this body of work by targeting the underlying localization problem directly, enabling global localization without external infrastructure, and doing so in a semantically dynamic environment representative of those faced by both assistive and service robots.

In contrast to prior methods that treat semantics as static landmarks, ShelfAware models them as statistical distributions over object counts and arrangements, enabling localization that is both robust to semantic flux and compatible with low-cost visual sensing. This probabilistic treatment of semantics situates ShelfAware within the broader context of semantic particle filtering, extending its applicability beyond assistive scenarios to general quasi-static domains.

\section{OUR APPROACH}
The objective of our proposed method is to achieve robust global localization in quasi-static indoor environments
These environments, which include warehouses, retail spaces, and laboratories, challenge conventional geometric localization methods due to visually repetitive structures, dynamic occlusions, and semantic drift \cite{gharpureRobotassistedShoppingBlind2008}. ShelfAware addresses these challenges by fusing geometric and semantic cues within a probabilistic particle filtering framework tailored for vision-based sensing, and deployable on wearable or compact platforms that preclude LiDAR or wheel odometry.

\subsection{Semantic Particle Filter Overview}
\label{sec:shelfaware_overview}

\begin{figure}[t]
  \includegraphics[width=\columnwidth,trim=18 20 18 20,clip]{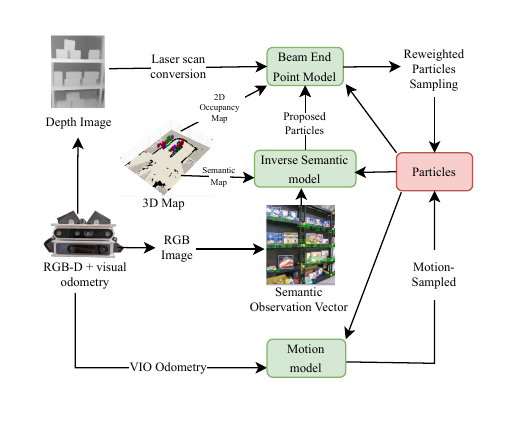}
  \caption{Data-flow diagram for ShelfAware. The semantic particle filter fuses depth likelihood with semantic likelihood and uses an inverse semantic model to propose high-quality particles for global localization and recovery.}
  \label{fig:architecture}
\end{figure}

ShelfAware builds on the MCL framework \cite{foxKLDSamplingAdaptiveParticle2001}, augmenting standard depth likelihoods with a probabilistic semantic observation model that remains informative under semantic variability. Each particle represents a pose hypothesis $\mathbf{x}_t^{(i)}$ with weight $\mathbf{w}_t^{(i)}$, updated via motion, geometric (depth), and semantic observations (Fig. ~\ref{fig:architecture}). 

Unlike approaches that treat detected objects as fixed landmarks, ShelfAware models the semantic state of the environment as distributions over class counts and coarse spatial arrangement. At runtime, the system forms a compact semantic observation vector from the live RGB-D frame and compares it to expected semantic signatures derived from a hybrid map (Sec.~\ref{sec:shelfaware_semantic_mapping}). The resulting semantic similarity acts both as (i) a \emph{forward} observation likelihood for weighting particles and (ii) a query metric for an \emph{inverse} semantic model that proposes high-quality pose hypotheses when global localization or recovery is needed (Sec.~\ref{sec:shelfaware_mcl}). 

The joint observation model factors geometric and semantic likelihoods as such:
\begin{equation}
  p(\mathbf{z}_t \mid \mathbf{x}_t, m) 
  = \eta \; p_d(\mathbf{z}_t^d \mid \mathbf{x}_t, m_d)\; p_s(\mathbf{z}_t^s \mid \mathbf{x}_t, m_s),
  \label{eq:combined_observation_model}
\end{equation}
where $\eta$ is a normalizer, $m_d$ and $m_s$ are the geometric and semantic maps respectively, $\mathbf{z}_t = (\mathbf{z}_t^d, \mathbf{z}_t^s)$ are the depth and semantic observations (Secs. \ref{sec:shelfaware_semantic_obs_model},~\ref{sec:shelfaware_depth_obs_model}). \added{This factorization assumes conditional independence: given pose $\mathbf{x}_t$, depth $\mathbf{z}_t^d$ and semantic $\mathbf{z}_t^s$ measurements have independent noise.}

\subsection{Semantic Mapping}
\label{sec:shelfaware_semantic_mapping}

ShelfAware maintains a hybrid map combining geometric structure with semantic information. First, we construct a standard 2D occupancy grid map of traversable space using GMapping \cite{grisetti2007improved}, with $10 \times 10$~cm resolution. This resolution supports precise ray casting for the depth observation model within the MCL framework and aligns with the accuracy needed for potential downstream manipulation tasks \cite{agrawalShelfHelpEmpoweringHumans2023}.

Second, we build a semantic map overlaid on the occupancy grid. The semantic map discretizes the volume into $20 \times 20$~cm cells in $(x,y)$ and $30$~cm in $z$ to balance computational efficiency with the physical scale of objects and the environment. Each voxel stores a running distribution over observed object \emph{classes} and their \emph{counts}, yielding a coarse, distributional representation that captures typical arrangements without assuming fixed landmark identities.

\added{To demonstrate that our system can work with different vision pipelines, we populate this semantic map using two distinct pipelines, depending on the evaluation environment:}

\added{\paragraph{Pipeline A: Supervised Closed-Set Semantics} For rigorous quantitative benchmarking in a controlled mock environment (Sec. IV-A) with fixed and known number of classes, we utilize a YOLOv9 detector fine-tuned on the SKU-110K dataset \cite{goldman2019precise} to extract generic ``shelf item" bounding boxes. A ResNet50-based classifier then assigns these detections to one of $C=14$ manually defined application-level classes.}

\begin{figure}[t]
    \centering
  \includegraphics{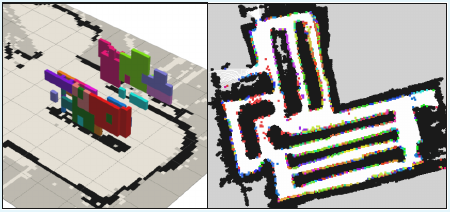}
  \caption{The semantic map overlaid on the occupancy grid. Each voxel stores a distribution over object class counts. Ray casting on this semantic layer yields the expected semantic vector $\mathbf{v}_{\text{sem}}$ comprising class counts, distances, and angles. \textit{Left:} Mock store, \textit{Right:} Real Store.}
  \label{fig:sem_map}
\end{figure}

\added{\paragraph{Pipeline B: Self-Supervised Open-Set Semantics} To demonstrate zero-shot scalability for real-world deployment in a 3,500 sq. ft. operational grocery store, we eliminate manual annotation entirely. Using the same class-agnostic detector, we extract a 768-dimensional vector for each bounding box crop via a pre-trained DINOv3 Vision Transformer. We then cluster these features using K-Means, automatically defining $C=60$ (obtained via grid-search) semantic categories (Fig. \ref{fig:sem_map}:right).}

\added{\paragraph{3D Projection and Map Integration}}
\added{Regardless of the chosen perception pipeline, detected} objects are projected into the map frame using the RGB-D depth channel and camera calibration.


Given pixel coordinates $\tilde{\bm{u}} = [u, v, 1]^\top$ at the bounding box's \emph{median} depth $Z_c$, the 3D world position is $\bm{X}_w = \bm{t}_{wc} + Z_c \bm{R}_{wc} \bm{K}^{-1} \tilde{\bm{u}}$, where $\bm{K}$ is the camera intrinsic matrix, and $\bm{R}_{wc}, \bm{t}_{wc}$ denote the camera-to-world transform.

To mitigate depth noise and vision errors, we refine each estimated product position along the camera ray using Bresenham’s line algorithm \cite{koopman1987bresenham} until it aligns with the nearest occupied occupancy map cell, preventing minor misalignments.
Figure~\ref{fig:sem_map} visualizes the resulting semantic layer overlaid on the occupancy grid; for clarity, the plot shows only the dominant class per cell, while the map stores full per-class count distributions.

\subsection{Semantic Observation Vector and Semantic Similarity}
\label{sec:shelfaware_semantic_obs_model}

\begin{figure}[t]
  \centering
  \includegraphics[width=0.9\columnwidth]{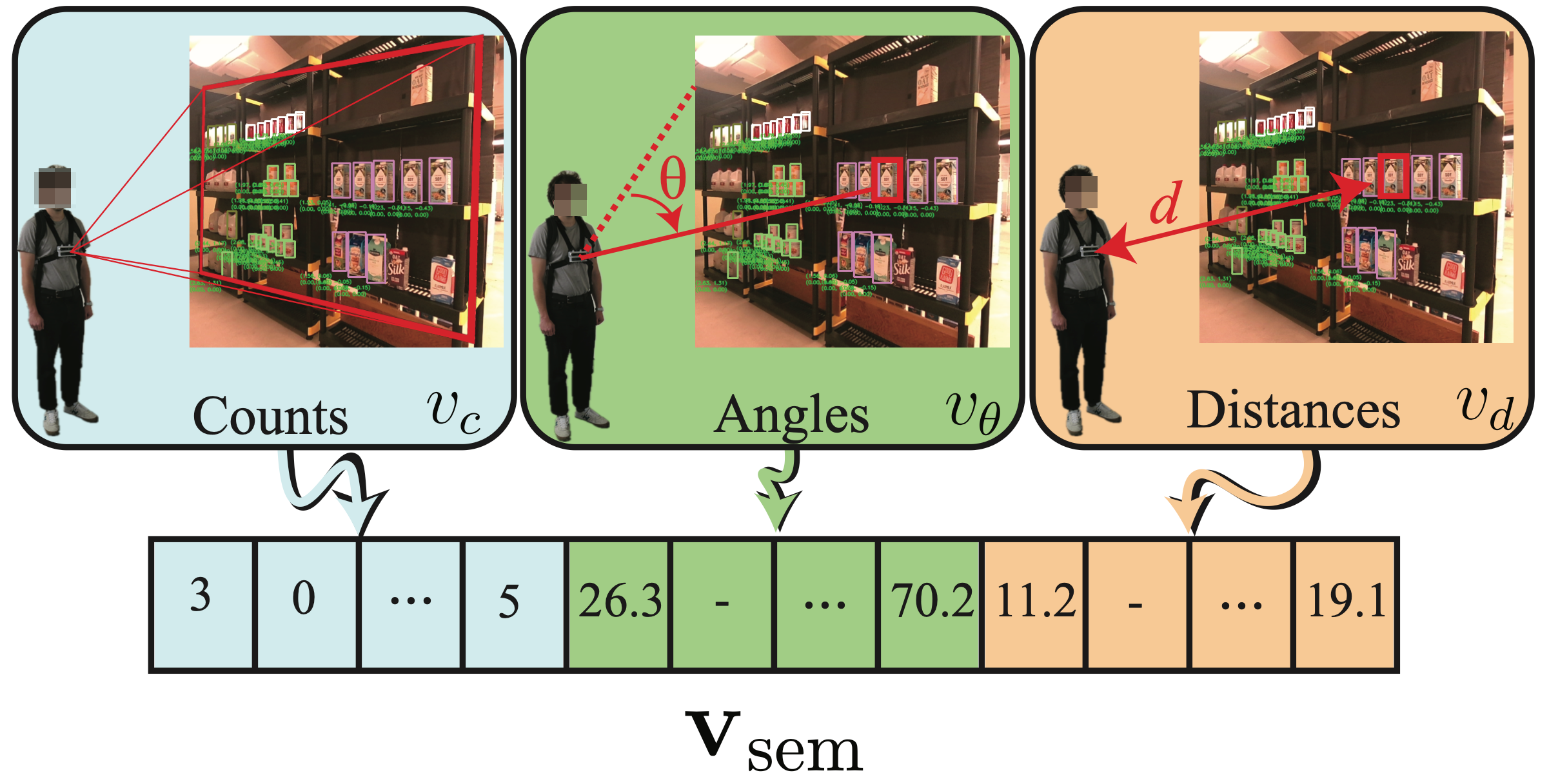}
  \caption{Semantic vector $\mathbf{v}_{\text{sem}} = [\mathbf{v}_c, \mathbf{v}_\theta, \mathbf{v}_d]$. The count vector $\mathbf{v}_c$ captures the number of items detected in each class; $\mathbf{v}_\theta$ and $\mathbf{v}_d$ capture mean relative bearings and ranges for those.}
  \label{fig:vector}
\end{figure}

ShelfAware generates a semantic observation $\mathbf{z}_t^s$ from the live camera view (Fig.~\ref{fig:vector}). This observation concatenates three sub-vectors:
(i) a class-count vector $\mathbf{v}_c$ (what is present), 
(ii) a mean range vector $\mathbf{v}_d$ (how far), and 
(iii) a mean bearing vector $\mathbf{v}_\theta$ (in which direction), 
each aggregated over detections for the classes visible at time $t$.

To compare live and expected semantic observations, we define a composite similarity
\begin{equation}
    S(\mathbf{z}^s, \hat{\mathbf{z}}^s) = \alpha S_{\text{counts}} + \beta S_{\text{distance}} + \gamma S_{\text{angle}},
    \label{eq:semantic_similarity}
\end{equation}
with $\alpha,\beta,\gamma \ge 0$ and $\alpha+\beta+\gamma=1$. This score is used both as (i) a forward semantic likelihood $p_s(\mathbf{z}_t^s \mid \mathbf{x}_t, m_s)$ in \eqref{eq:combined_observation_model} and (ii) a query metric for the inverse model (Sec.~\ref{sec:shelfaware_mcl}).

\paragraph{Counts.}
We normalize $\mathbf{v}_c$ to form a categorical distribution over observed classes and compare it with the expected distribution via Jensen–Shannon divergence (JSD). Defining $P$ and $Q$ as the normalized count distributions and $M=\frac{1}{2}(P+Q)$, we set $S_{\text{counts}} = 1-\mathrm{JSD}$ with $\mathrm{JSD}(P\parallel Q) = \sqrt{\tfrac{1}{2} \sum_{i=1}^{C} \left( P(i) \log \tfrac{P(i)}{M(i)} + Q(i) \log \tfrac{Q(i)}{M(i)} \right)}$. This choice is symmetric, bounded, and robust to sparsity in per-frame detections. \added{By comparing normalized distributions rather than absolute counts, this JSD formulation provides inherent robustness to perception failures. Transient false negatives (e.g., missed bounding boxes due to motion blur) or false positives do not catastrophically invalidate the observation, but rather smoothly degrade the similarity score, preventing abrupt particle depletion.}

\paragraph{Distances and Angles.}
The spatial components $\mathbf{v}_d$ and $\mathbf{v}_\theta$ are compared using L2 distances. For unbounded ranges, we map the L2 error $d_{\text{distance}}$ to $S_{\text{distance}}=1/(1+d_{\text{distance}})$. For angles, which lie within the camera field-of-view (FOV), we use $S_{\text{angle}} = 1 - (d_{\text{angle}}/\mathrm{FOV})$. 

\subsection{Depth Observation Model}
\label{sec:shelfaware_depth_obs_model}

To incorporate geometric information, we synthesize a 2D laser scan from a central horizontal band of the depth image. The depth observation at time $t$ is $\mathbf{z}_t^d = \{ z_t^{d,(k)} \}_{k=1}^K$. Given the expected range $\hat{z}^{(k)}(\mathbf{x}_t,m_d)$ for the $k$-th beam (via occupancy-map ray casting), we use the beam end-point mixture \cite{foxKLDSamplingAdaptiveParticle2001}:
\begin{align}
&p_{\mathrm{hit}}(z_t^{d,(k)} \mid \mathbf{x}_t, m_d) = \mathcal{N}\!\big(z_t^{d,(k)};\,\hat{z}^{(k)}(\mathbf{x}_t, m_d), \sigma_{\mathrm{hit}}^2\big), \label{eq:beam-hit} \\
&p_{\mathrm{short}}(z_t^{d,(k)} \mid \mathbf{x}_t, m_d) = \lambda e^{-\lambda z_t^{d,(k)}}\,\mathbf{1}_{[0,\hat{z}^{(k)}(\mathbf{x}_t,m_d)]}(z_t^{d,(k)}), \label{eq:beam-short} \\
&p_{\mathrm{max}}(z_t^{d,(k)}) = \delta(z_t^{d,(k)}-z_{\max}), \quad p_{\mathrm{rand}}(z_t^{d,(k)}) = 1/z_{\max}, \label{eq:beam-max-rand}
\end{align}
\begin{equation}
p_d(z_t^{d,(k)} \mid \mathbf{x}_t, m_d) = w_{\mathrm{h}}\,p_{\mathrm{hit}} + w_{\mathrm{s}}\,p_{\mathrm{short}} + w_{\mathrm{m}}\,p_{\mathrm{max}} + w_{\mathrm{r}}\,p_{\mathrm{rand}},
\label{eq:beam-mixture-split}
\end{equation}

where $m_d$ represents the 2D occupancy map, $z_{\max}=6.0\text{ m}$ is the maximum reliable sensor range, $\sigma_{\mathrm{hit}}=2.0$ is the standard deviation of the expected depth, and $\lambda=0.1$ is the intrinsic decay parameter for short readings. The empirically tuned mixture weights ($w_{\mathrm{h}}=0.85$, $w_{\mathrm{s}}=0.05$, $w_{\mathrm{m}}=0.05$, $w_{\mathrm{r}}=0.05$) define the relative probability of a valid hit, a short reading, a max-range reading, and a random noise reading, respectively, and are constrained to sum to one. Assuming conditional independence across beams, the joint likelihood for $\mathbf{z}_t^d$ is
\begin{equation}
p_d(\mathbf{z}_t^d \mid \mathbf{x}_t, m_d) = \prod_{k=1}^{K} p_d\!\big(z_t^{d,(k)} \mid \mathbf{x}_t, m_d\big).
\label{eq:depth-band-likelihood}
\end{equation}
We compute $\hat{z}^{(k)}(\mathbf{x}_t, m_d)$ efficiently using CDDT-accelerated ray casting \cite{walsh2018cddt}, which runs on the CPU and leaves GPU resources available for the semantic perception pipeline.

\subsection{Localization with an Inverse Semantic Model}
\label{sec:shelfaware_mcl}

ShelfAware integrates the depth and semantic models within an MCL filter, maintaining weighted particles $\{(\bm{x}_t^{(i)}, \bm{w}_t^{(i)})\}_{i=1}^N$ over the pose belief. The method’s key innovation is an \emph{inverse semantic model} that proposes high-quality pose hypotheses directly from live semantic observations, enabling fast \emph{global localization} and recovery from tracking failures without special-case handling.

\paragraph{Offline Pre-computation}
We precompute expected semantic observations $\hat{\bm{z}}^s(\bm{x}, m_s)$ for discretized poses $\bm{x}=(x,y,\theta)$ over the free space (10cm cells; 36 orientation bins). For each pose, we ray cast the 3D semantic map (Sec.~\ref{sec:shelfaware_semantic_mapping}) to obtain the expected semantic vector. All vectors are cached in a hashmap (76\,MB for our store environment). We also build a reverse index \texttt{class\_to\_poses} (2.1\,MB) mapping each product class to the set of poses from which that class is visible, enabling efficient candidate pruning at runtime.

\paragraph{Online Localization Loop.}
Algorithm~\ref{alg:semantic_mcl} summarizes the online filter. Particles are propagated using VIO-based motion. Each iteration forms a live semantic vector $\bm{z}_t^s$ and estimates an expected semantic view $\hat{\bm{z}}_t^s$ at the current pose estimate $\hat{\bm{x}}$. A semantic consistency check compares $S(\bm{z}_t^s,\hat{\bm{z}}_t^s)$ against a threshold and verifies that the count subvector has sufficient mass ($\lVert \bm{z}_t^{s,\text{count}} \rVert_1 > \tau_\kappa$). If the check fails, we query the inverse model: (i) use \texttt{class\_to\_poses} to form a candidate set (union over observed classes), (ii) score candidates by $S$ (Eq.~\ref{eq:semantic_similarity}), (iii) inject particles at the top-$k$ poses, and (iv) reweight the entire set with both depth (Sec.~\ref{sec:shelfaware_depth_obs_model}) and forward semantic likelihoods (Eq.~\ref{eq:combined_observation_model}). Otherwise, we update using only the depth likelihood for efficiency. This procedure enables rapid convergence from unknown initial conditions and robust recovery from drift.

\begin{algorithm}
\caption{Online Localization with ShelfAware}
\label{alg:semantic_mcl}
\begin{algorithmic}[1]
\Statex \hspace{-0.6cm} \textbf{Input:}
\quad Sensor data: image $i_t$, depth $d_t$, odometry $o_t$
\Statex \quad Pre-computed map data: semantic vectors $\hat{\mathbf{z}}^s(\mathbf{x})$,
\Statex \hspace{-0.6cm} \textbf{Output:} Updated particle set $P_{t+1}$
\Statex \hspace{-0.6cm} \textbf{Initialize:} Particle set $P_{t}$ sampled uniformly over free space
\While{True}
  \State $P_t \gets \text{MotionModel}(P_t, o_t)$
  \State $z^s_t \gets \text{GenerateSemanticVector}(i_t, d_t)$
  \State $\hat{x} \gets \text{ExpectedPose}(P_t, W_t)$
  \State $\hat{z}^s_t \gets \text{CalculateExpectedSemanticObs}(\hat{x})$
  \State \textit{// Semantic consistency and information sufficiency}
  \State $\text{sim} \gets \text{CalculateSemanticSimilarity}(z^s_t, \hat{z}^s_t)$
  \If{$\text{sim} < \tau_{\text{sim}}$ \textbf{and} $\lVert z^{s, \text{count}}_t \rVert_1 > \tau_{\kappa}$}
        \State $C \gets \text{GetObservedClasses}(\mathbf{z}_t^s)$
        \State $\text{Candidates} \gets \bigcup_{c \in C} \texttt{class\_to\_poses}[c]$
        \State $\text{Scores} \gets [S(\mathbf{z}_t^s, \hat{\mathbf{z}}^s(\mathbf{x})) \text{ for } \mathbf{x} \in \text{Candidates}]$
        \State $\text{TopPoses} \gets \text{GetTopK}(\text{Candidates}, \text{Scores}, k)$
        \State $P_t \gets \text{InjectParticles}(P_t, \text{TopPoses})$
        \State $W_{\text{depth}} \gets \text{DepthObservationModel}(P_t, d_t)$
        \State $W_{\text{sem}} \gets \text{SemanticObservationModel}(P_t, i_t, d_t)$
        \State $W_t \gets W_{\text{depth}} \odot W_{\text{sem}}$ \Comment{element-wise product}
  \Else
    \State $W_t \gets \text{DepthObservationModel}(P_t, d_t)$
  \EndIf
  \State $W_t \gets \text{NormalizeWeights}(W_t)$
  \State $P_t \gets \text{Resample}(P_t)$
  \State $P_{t+1} \gets P_t$
\EndWhile
\end{algorithmic}
\end{algorithm}

\subsection{Hardware and Form Factors}
\label{sec:shelfaware_hardware}

\begin{figure}[t]
\centering
  \includegraphics[width=\columnwidth]
  {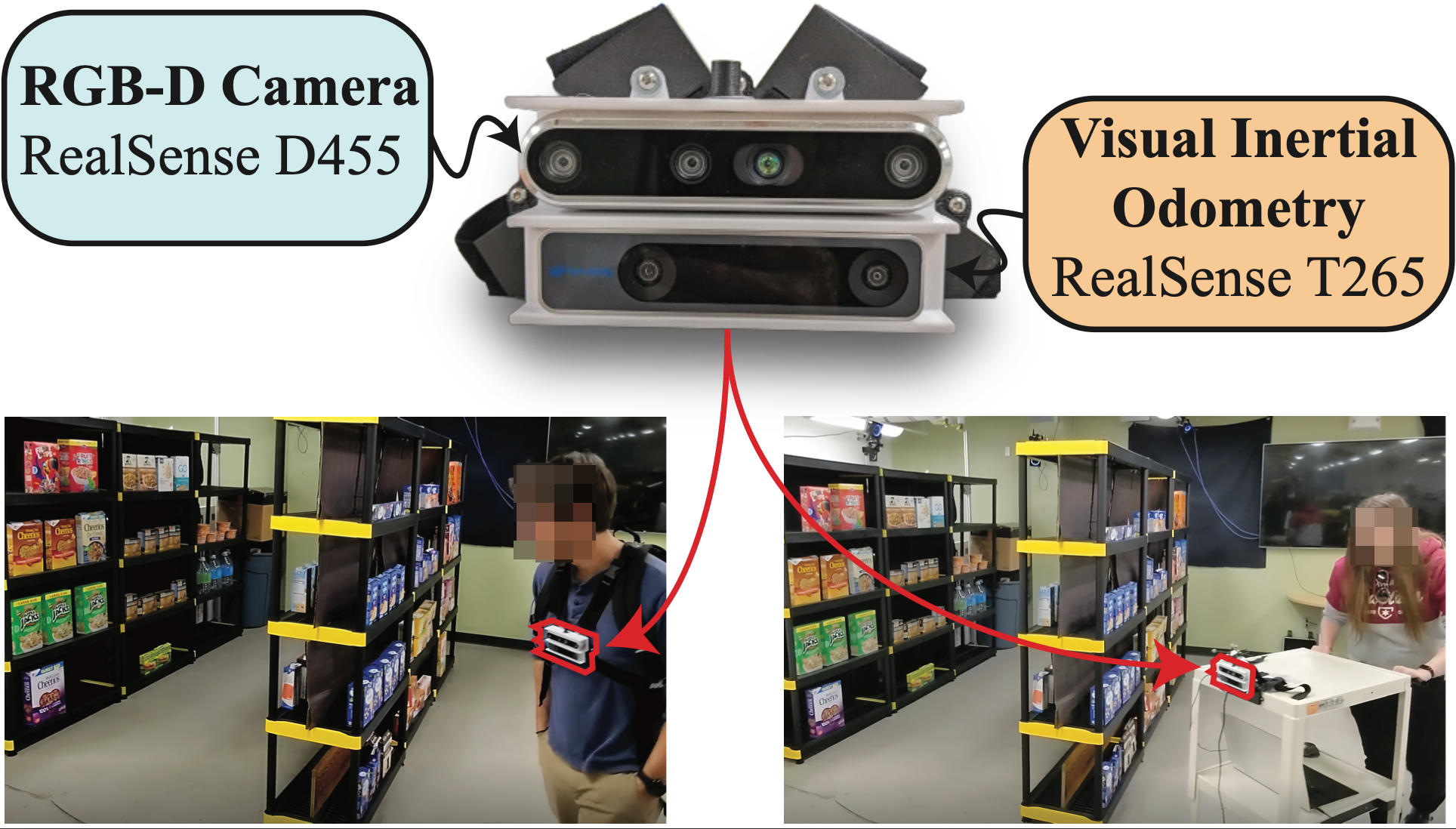}
  
  \caption{ShelfAware hardware. A lightweight two-camera system with a 3D-printed mount was used throughout our experiments (top). This design allowed evaluation across a wearable chest mount (left) and a cart-mounted setup (right).}
  \label{fig:hardware}
\end{figure}

To meet the constraints of low-profile, wearable, or cart-mounted platforms, ShelfAware uses compact vision sensors: (i) an Intel RealSense D455 RGB-D camera ($\sim$103\,g) for color and depth, and (ii) an Intel RealSense T265 VIO camera ($\sim$60\,g) for odometry. Both are housed in a single 3D-printed mount and connected via USB to a Dell G15 laptop. This configuration supports two form factors: a chest-mounted wearable (laptop in a backpack) and a cart-mounted setup (Fig.~\ref{fig:hardware}). As discussed in Secs.~\ref{sec:shelfaware_semantic_mapping}--\ref{sec:shelfaware_mcl}, the system exploits CPU-based CDDT ray casting \cite{walsh2018cddt} to reserve GPU resources for real-time semantic perception, enabling practical deployment without LiDAR or wheel encoders.

\section{Experimental Evaluations}


\added{We evaluate ShelfAware’s ability to perform reliable, real-time global localization in quasi-static, semantically dense, and GPS-denied environments using compact vision sensors. Specifically, we assess its robustness to dynamic occlusions and sparse semantics, and its form factor sensitivity across wearable and cart-mounted configurations. We adopt these form factors to reflect practical compute constraints and to stress-test the method under depth-camera noise (lower scan frequency/point density than LiDAR) \cite{liu2007effect}, intentionally avoiding the usability barriers of handheld devices \cite{kim2013usability} and the social stigma of head-mounted systems \cite{lee2020pedestrian}. Ultimately, these evaluation goals reflect the start-anytime and recover-anytime operation demanded by practical deployments and shared-control use cases \cite{zhang2023follower,kamikubo2025beyond,arora2024visionai}.}

\subsection{Experimental Setup}
We conduct our \added{controlled} evaluation in a mock grocery store stocked with 150 products grouped into 14 object categories across nine shelves and three aisles. All experiments ran on a Dell G15 laptop (Intel Core i7-11800H (2.3 GHz); NVIDIA RTX~3060, 6GB VRAM; 32GB RAM) using the same vision-only sensor suite described in Sec.~\ref{sec:shelfaware_hardware}: an Intel RealSense D455 RGB-D camera and a RealSense T265 VIO camera (Fig.~\ref{fig:hardware}). Ground-truth 2D poses were obtained using an OptiTrack motion-capture system. We aligned the OptiTrack frame to the map frame offline by time-synchronizing mapping trajectories and solving for the rigid transform via RANSAC \cite{derpanis2010overview}, enabling direct comparison of estimated and true poses. RGB-D streams were recorded at 30Hz and VIO at 200Hz (from the T265 IMU). This evaluation setting stresses depth-camera uncertainties that degrade purely geometric localization methods \cite{liu2007effect}. 

We considered four conditions to measure and characterize the robustness of the proposed method:
\begin{enumerate}
    \item \textbf{Cart:} Cameras mounted on a cart (Fig.\ref{fig:hardware}-right) \cite{nicholson2009shoptalk}.
    \item \textbf{Wearable:}  Cameras chest-mounted; laptop carried in a backpack (Fig.\ref{fig:hardware}-left), stressing odometry noise introduced by gait.
    \item \textbf{Dynamic Obstacles:} A person intermittently walked in front of the cameras to occlude vision.
    \item \textbf{Sparse Semantics:} To mimic depleted stock, we randomly removed 25\% and 50\% of products from shelves, reducing category signal.
\end{enumerate}


We collected 5 trajectory sequences per condition (20 total, S1--S20), each $\sim40$\,s in duration. Between mapping and localization runs, we perturbed 20\% of shelf contents to induce semantic drift. Unless stated otherwise, particles were initialized \emph{uniformly over free space} to require true global localization \cite{kulyukinRoboCartRobotassistedNavigation2005,nicholson2009shoptalk}.


\subsection{Perception Pipeline Configurations}
We utilize the two perception pipelines introduced in Sec.~\ref{sec:shelfaware_semantic_mapping}. For the mock store (Pipeline A), a YOLOv9 detector (fine-tuned on the SKU-110K dataset \cite{goldman2019precise}) proposes generic bounding boxes, which a ResNet50 classifier maps into 14 object categories (trained on 5,699 environment-specific frames). Grouping thousands of underlying SKU types into a small, fixed category set provides robustness against local inventory churn. Conversely, for the real-world operational store (Pipeline B), we combine YOLOv9's class-agnostic bounding box proposals with a zero-shot DINOv3 Vision Transformer and K-Means clustering. This open-vocabulary approach automatically extracts 60 semantic categories from the generic detections in a purely self-supervised manner, requiring zero environment-specific training or manual annotation to adapt to new inventory.

\subsection{Baselines and Metrics}
\textbf{Baselines.} We compare against \added{three methods:} (i) the standard ROS implementation of AMCL \cite{nav2_amcl}\added{,} (ii) an ablated ShelfAware variant that removes the semantic likelihood, yielding a pure \added{geometric} MCL baseline \cite{foxKLDSamplingAdaptiveParticle2001}\added{, and (iii) a Fixed Quantity Landmark (FQL) baseline. The FQL baseline represents standard explicit semantic localization approaches \cite{zimmermanRobustOnboardLocalization2022, zimmermanLongTermLocalizationUsing2023} by treating detected object classes as having fixed, deterministic counts.} For fairness, all methods used \textbf{1,500 particles}, identical VIO odometry and RGB-D stream.

\noindent\textbf{Metrics.} Following established conventions in the literature, we evaluate ShelfAware via:
\begin{itemize}
    \item \textbf{Global localization success.} A trial is a success if localization convergence occurs within the first 95\% of the trajectory and remains converged until the end.
    \item \textbf{Convergence time (s).} Time from start to the beginning of the final convergence for successful trials.
    \item \textbf{Tracking ATE.} Absolute Trajectory Error (translation/rotation RMSE) computed after final convergence until end-of-sequence.
\end{itemize}

\noindent\textbf{Convergence criterion.} \added{Anthropometric studies indicate a standing reach of $\sim$0.7--0.9 m \cite{sengupta2000maximum}; thus, localization errors below 0.7 m ensure targets remain within the reachable workspace.} We declare convergence when the estimated pose is within $0.7$~m translation and $\pi/4$ rad rotation of ground truth and stays within that threshold. 
We tuned semantic similarity weights by grid search and report results for $\alpha=0.4, \beta=0.4, \gamma=0.2$ (Sec.\ref{sec:shelfaware_semantic_obs_model}).


\begin{table*}[t]
\centering
\caption{Global localization results across setups in the mock store environment. Success is \% of 25 trials per setup. The convergence time and RMSE is calculated only for successful convergences.}
\label{tab:global}
\setlength{\tabcolsep}{3pt}\renewcommand{\arraystretch}{1.2}
\resizebox{\textwidth}{!}{%
\begin{tabular}{l|ccc|ccc|ccc|ccc|c}
\hline
 & \multicolumn{3}{c|}{Cart} & \multicolumn{3}{c|}{Wearable} & \multicolumn{3}{c|}{Dynamic} & \multicolumn{3}{c|}{Degraded/Sparse} & \multirow{2}{*}{\textbf{Overall G\%}} \\
\cline{2-13}
Method & Success & Time (s) & RMSE (m/rad) & Success & Time (s) & RMSE (m/rad) & Success & Time (s) & RMSE (m/rad) & Success & Time (s) & RMSE (m/rad) & \\
\hline
MCL        & 16\% & 1.06 & 0.41/0.27 & 12\% & 0.13 & 0.37/0.49 & 16\% & 0.13 & 0.25/0.04 & 16\% & 0.05 & 0.36/0.26 & 15\% \\
AMCL       & 20\%  & 17.41& \textbf{0.18}/\textbf{0.02} & 0\%  & - & - & 0\%  & - & - & 20\% & 3.36 & 0.34/\textbf{0.05} & 10\% \\
FQL        & 96\% & 1.98 & 0.38/0.20 & 96\% & 4.14 & \textbf{0.42/0.30} & 80\% & 2.75 & \textbf{0.36}/0.22 & 88\% & 4.07 & 0.42/\textbf{0.23} & 90\% \\
ShelfAware & \textbf{100\%} & \textbf{0.27} & 0.37/0.25 & \textbf{100\%} & \textbf{1.96} & 0.36/0.42 & \textbf{100\%} & \textbf{0.29} & 0.39/\textbf{0.15} & \textbf{88\%} & \textbf{1.83} & \textbf{0.31}/0.34 & \textbf{97\%} \\
\hline
\end{tabular}}
\end{table*}

\begin{table}[!ht]
\centering
\caption{Tracking Success Rate (T\%) in the mock store.}
\label{tab:tracking-success}
\setlength{\tabcolsep}{4pt}
\renewcommand{\arraystretch}{1.15}
\begin{tabular}{lcccccc}
\toprule
\textbf{Method} & \textbf{Cart} & \textbf{Wearable} & \textbf{Dynamic} & \textbf{Sparse} & \textbf{All} & \textbf{T-RMSE} \\
\midrule
MCL & 8\% & 8\% & 12\% & 12\% & 10\% & 0.27/0.13 \\
AMCL & 17\% & 0\% & 0\% & 20\% & 10\% & \textbf{0.25}/0.14 \\
FQL & \textbf{60}\% & 60\% & 68\% & 48\% & 59\% & 0.32/0.10 \\
\textbf{ShelfAware} & \textbf{60\%} & \textbf{72\%} & \textbf{80\%} & \textbf{52\%} & \textbf{66\%} & 0.29/\textbf{0.09} \\
\bottomrule
\end{tabular}
\end{table}

\subsection{Results: Global Localization}
Table \ref{tab:global} (four conditions, 25 trials each) summarizes success, convergence time, and RMSE. ShelfAware achieved a \textbf{97\%} overall success rate across 100 trials, versus \textbf{15\%} for MCL and \textbf{10\%} for AMCL. The mean time-to-convergence across all setups was \textbf{1.07s}. ShelfAware obtained the lowest translational RMSE in three out of four conditions. 

\begin{figure}[h]
  \centering
  \includegraphics[width=\columnwidth]{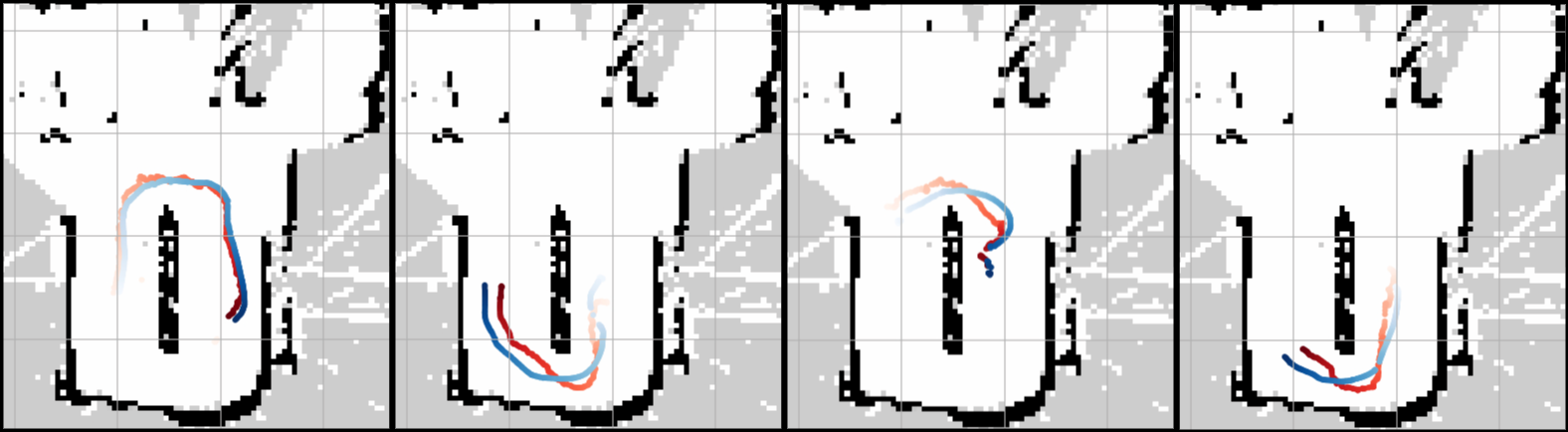}
  \caption{Examples of ground truth pose in blue and our estimated pose in red. Lighter to darker denotes the temporal progression of the trajectories. Each grid cell is $2$m$\times2$m.}
  \label{fig:trajectories}
\end{figure}

\subsection{Results: Tracking and Robustness}
To evaluate continuous tracking performance, we measure the Tracking Success Rate (T\%), defined as the percentage of trials that converge and remain strictly within the convergence thresholds until the end. Table \ref{tab:tracking-success} reports the T\% and the average tracking RMSE (T-RMSE) across all conditions. 

ShelfAware achieved an overall tracking success rate of \textbf{66\%}, outperforming the deterministic FQL baseline (59\%) and vastly exceeding the purely geometric MCL and AMCL methods (10\%). Importantly, the system demonstrated particular resilience against the two primary environmental stressors: dynamic occlusions and sparse semantics. Qualitative trajectories (Fig.\ref{fig:trajectories}) show rapid correction in repetitive aisles and robustness to occlusion bursts. These results mirror the global-localization advantage, demonstrating that probabilistic semantic tie-breaking maintains spatial consistency despite active dynamic obstacles and severe stock depletion.


The full pipeline ran at 9.6Hz on a standard consumer laptop with a mid-tier GPU, achieving real-time throughput without LIDAR or wheel encoders.

\subsection{Real-World Open-Vocabulary Evaluation}
\label{sec:real_store_eval}


\added{To demonstrate ShelfAware's scalability and its independence from closed-set object detectors, we deployed the open-vocabulary pipeline (Pipeline B; Sec.~\ref{sec:shelfaware_semantic_mapping}) using the wearable form factor in a 3,500 sq.\ ft.\ operational specialty grocery store, generating a highly tractable 380 MB semantic map (Fig.~\ref{fig:sem_map}, right). Because deploying a motion capture system was infeasible in a public retail space, ground truth was established via optimized RTAB-Map SLAM trajectories from the mapping phase. From this, we sampled 10 random evaluation trajectories ($\sim$40 seconds each at a normal walking pace). To ensure a fair comparison, we conducted a grid search over filter parameters for all baselines, reporting the highest-performing configurations.

Table \ref{tab:real-store} presents these results. In this large, geometrically aliased environment, purely geometric methods (MCL, AMCL) collapse ($\le 8\%$ success). Using the semantic vision pipeline alongside depth observations, ShelfAware successfully achieved global localization in 62\% of trials, significantly outperforming all baselines. We note that ShelfAware requires more time to achieve initial global convergence in this setting than the FQL baseline. This is an expected computational trade-off: to achieve its higher success rate across a 3,500 sq. ft. space, ShelfAware's inverse semantic model must evaluate and inject particle proposals from a vastly larger candidate pose space.

We also performed an upper-bound ablation using ground-truth (GT) semantic vectors (obtained via raycasting the semantic map from the known RTAB-Map poses) while disabling the noisy depth likelihood. Under these perfect semantic conditions, ShelfAware achieved 98\% global localization success and 92\% tracking stability. In contrast, the FQL baseline only achieved 82\% and 70\% under the same perfect-vision conditions. This ablation proves that our core methodological contribution of modeling semantics as continuous probability distributions over object counts, inherently extracts more robust localization signals than fixed-quantity representations, even when perception is flawless. Furthermore, the ablation revealed that while depth observations are necessary to stabilize the noisy visual features in the ``real'' condition, incorporating depth actually degrades performance when semantic perception is perfect, highlighting the distinct aliasing risks of depth sensors in geometrically ambiguous domains.}

\begin{table}[t]
\centering
\caption{Real-world grocery store evaluation. ``Upper-Bound Semantics" isolates the filter using ground-truth semantics, while ``Real Vision" uses the live RGB image. Metrics include Global Localization Success (G\%), Tracking Success (T\%), Time to Global Convergence (G-Time), and translational/rotational RMSE (m/rad).}
\label{tab:real-store}
\setlength{\tabcolsep}{3pt}
\renewcommand{\arraystretch}{1.15}
\begin{tabular}{lccccc}
\toprule
\textbf{Method} & \textbf{G\%} & \textbf{T\%} & \textbf{G-RMSE} & \textbf{T-RMSE} & \textbf{G-Time(s)} \\
\midrule
MCL & 8\% & 6\% & \textbf{0.30}/0.25 & \textbf{0.31}/0.15 & 17.4 \\
AMCL & 6\% & 4\% & 0.36/0.13 & 0.41/\textbf{0.03} & 4.1 \\
\multicolumn{6}{c}{\textit{Real Vision (RGB + Depth)}} \\
FQL & 42\% & 38\% & 0.41/\textbf{0.09} & 0.37/0.06 & \textbf{2.9} \\
\textbf{ShelfAware} & \textbf{62\%} & \textbf{40\%} & 0.34/0.12 & \textbf{0.31}/0.10 & 6.2 \\
\midrule
\multicolumn{6}{c}{\textit{Upper-Bound Semantics (GT Semantics, No Depth)}} \\
FQL & 82\% & 70\% & 0.41/0.31 & \textbf{0.14}/0.07 & \textbf{1.6} \\
\textbf{ShelfAware} & \textbf{98\%} & \textbf{92\%} & \textbf{0.35/0.27} & 0.19/\textbf{0.06} & 3.2 \\
\bottomrule
\end{tabular}
\end{table}

\subsection{Discussion and Limitations}
\paragraph*{Implications for assistive devices for People with Visual Impairment (PVI)}
Although our core contribution is a general method for vision-based localization in quasi-static environments, the results have direct implications for assistive navigation. First, \emph{start-anytime} operation is crucial in shared-control assistive use: users often want to invoke assistance on demand rather than run a fully autonomous device continuously. ShelfAware’s rapid global localization (mean $1.07$s across scenarios; Table \ref{tab:global}) and its ability to recover from lost tracking (Table \ref{tab:tracking-success}) align with these constraints by enabling on-demand pose estimation and re-localization without external infrastructure. Second, our chosen form factors reflect evidence on acceptance and usability: cane-mounted sensors face handling barriers and are often undesired by users \cite{kim2013usability} and head-mounted systems are frequently rejected for bulk and stigma \cite{lee2020pedestrian}, while cart-mounted or less bulky wearable solutions have shown promise \cite{nicholson2009shoptalk}. These findings motivate the chest-mounted and cart configurations in Fig. \ref{fig:hardware} and Sec. \ref{sec:shelfaware_hardware}. Finally, a class-level semantic representation is a practical fit for retail navigation: detectors can see many \emph{classes}, but mapping detections into a compact set is both stable under SKU churn and sufficiently distinctive for localization, as evidenced by the high success rates in Table \ref{tab:global}.

\paragraph*{Path to an assistive navigation stack}
ShelfAware provides spatial grounding that upstream guidance and downstream interaction modules can exploit. In shopping contexts, prior work has focused on product retrieval or fine-grained identification near the correct shelf \cite{nicholson2009shoptalk,agrawalShelfHelpEmpoweringHumans2023}. Our results address the prerequisite of reliably \emph{reaching} the correct aisle/shelf in semantically dynamic, geometrically repetitive layouts. Integrating ShelfAware with wayfinding, obstacle avoidance, and product-retrieval interfaces (e.g., speech or haptics) is a natural next step toward end-to-end assistive experiences. Importantly, because our approach requires only vision sensors and VIO, it avoids environmental augmentation (e.g., RFID/beacons) and aligns with infrastructure-free deployments \cite{kulyukinRoboCartRobotassistedNavigation2005,lopez-de-ipinaBlindShoppingEnablingAccessible2011,nicholson2009shoptalk,bigham2010vizwiz}.


\paragraph*{Limitations}
Four limitations warrant discussion. (i) \emph{Large-scale drift and map updates.} Our inverse proposal mechanism can overcome moderate semantic drift by re-injecting particles, provided sufficient semantic cues remain visible in the live scene. However, massive inventory reorganizations outstrip this capacity. ShelfAware currently lacks an automated update mechanism and significant changes require rebuilding the semantic map from scratch. (ii) \emph{Perception failures.} Prolonged occlusions, motion blur, or severe category sparsity can reduce the information mass in the semantic vector, delaying inverse proposals and weakening forward likelihoods. (iii) \emph{Computational complexity.} Precomputing the inverse semantic cache is strictly offline and scales linearly with the number of discrete states. Online, evaluating $K$ depth beams and $C$ semantic classes per particle scales linearly as $\mathcal{O}(N)$, since $K$ and $C$ are small constants. Recovering via particle injection takes only $\mathcal{O}(1)$ hash-map lookups.
(iv) \emph{Human-factors validation.} Our form-factor choices are informed by prior literature and informal feedback, but we did not conduct PVI user studies; measuring usability and trust with PVI participants, including multi-hour battery tests and interaction design, is essential future work. 

\section{Conclusion}

\added{In this work we propose \emph{ShelfAware}, a novel method to address the challenging problem of vision-based localization in \emph{quasi-static} indoor environments, where geometry is repetitive and local semantics evolve. ShelfAware models semantics as \emph{distributional evidence over object categories} and couples this representation with an inverse semantic proposal mechanism inside an MCL framework, enabling the filter to remain informative under semantic drift and to generate targeted global pose hypotheses when needed. 

Our core contributions are twofold: (i) a semantic mapping and observation design that encodes category counts and coarse spatial structure as probabilistic distributions; and (ii) a real-time particle filter that fuses depth likelihoods with this semantic similarity, utilizing precomputed semantic viewpoints for inverse pose proposals. We demonstrate the efficacy through extensive empirical evaluation, achieving robust \emph{global localization} and \emph{tracking} across both a highly controlled mock setup and a real-world grocery store.

In our mock environment, ShelfAware achieved \textbf{97\%} overall global-localization success across four conditions vastly outperforming geometric baselines (15\% MCL, 10\% AMCL) and deterministic Fixed Quantity Landmark baselines (90\% FQL). Furthermore, when deployed in a massive 3,500 sq. ft. retail store using an open-vocabulary vision pipeline, it successfully globally localized in \textbf{62\%} of trials (compared to FQL's 42\%), proving its scalability and robustness to severe perception noise. An upper-bound ablation revealed that ShelfAware extracts significantly more localization signal from perfect semantic data than deterministic methods when provided with ground-truth semantics, ShelfAware achieved \textbf{98\%} global success and \textbf{92\%} tracking success, compared to just \textbf{82\%} and \textbf{70\%} for the FQL.

Beyond retail, the method applies to service and mobile robots in warehouses and offices, where quasi-static semantics and ambiguous geometry are common. In assistive contexts, the ability to \emph{start anytime} and \emph{relocalize} quickly supports shared-control and infrastructure-free deployment. Future work includes larger-scale evaluations across diverse layouts, embedded implementations, and online maintenance of the semantic layer, further advancing practical, vision-only localization in dynamic real-world environments.}

\bibliographystyle{IEEEtran}
\bibliography{references}

\end{document}